\documentclass{sig-alternate}
\newfont{\mycrnotice}{ptmr8t at 7pt}
\newfont{\myconfname}{ptmri8t at 7pt}
\let\crnotice\mycrnotice%
\let\confname\myconfname%

\usepackage[german,american]{babel}
\usepackage[T1]{fontenc}
\usepackage[utf8]{inputenc}
\usepackage{times}
\usepackage{textcomp}
\usepackage{graphicx}
\usepackage{amssymb}
\usepackage{amsmath}
\usepackage{microtype} 
\usepackage{multirow}

\usepackage{url}
\renewcommand{\tt}{\fontencoding{OT1}\fontfamily{cmtt}\selectfont}

\usepackage[usenames,dvipsnames]{color}
\usepackage{booktabs}
\usepackage[numbers,sectionbib]{natbib}
\makeatletter
\def\@copyrightspace{\relax}
\makeatother

\begin{document}

\title{Ensemble of Neural Classifiers for Scoring Knowledge Base Triples}
\subtitle{The Lettuce Triple Scorer at WSDM Cup 2017}

\numberofauthors{3}
\author{
\alignauthor
Ikuya Yamada\\
\affaddr{Studio Ousia}\\
\affaddr{ikuya@ousia.jp}\\
\alignauthor
Motoki Sato\\
\affaddr{Nara Institute of Science and Technology}\\
\affaddr{sato.motoki.sa7@is.naist.jp}\\
\alignauthor
Hiroyuki Shindo\\
\affaddr{Nara Institute of Science and Technology}\\
\affaddr{shindo@is.naist.jp}\\
}

\maketitle

\begin{abstract}
This paper describes our approach for the triple scoring task at the WSDM Cup 2017.
The task required participants to assign a relevance score for each pair of entities and their types in a knowledge base in order to enhance the ranking results in entity retrieval tasks.
We propose an approach wherein the outputs of multiple neural network classifiers are combined using a supervised machine learning model.
The experimental results showed that our proposed method achieved the best performance in one out of three measures (i.e., Kendall's $\tau$), and performed competitively in the other two measures (i.e., accuracy and average score difference).
\end{abstract}

\section{Introduction}

In the last decade, huge online structured knowledge bases (KBs) such as Wikidata \cite{Vrandecic2014a}, Freebase \cite{Bollacker2008}, and DBpedia \cite{Auer2007} have emerged.
These KBs contain an enormous number of entities (e.g., people) and their types (e.g., professions and nationalities).\footnote{Entities and their types can be easily extracted from KB triples where their subjects refer to entities and their objects are the corresponding types. Here, the target triple is a triple describing a relation of which the object can be one among a limited set of values such as the nationalities of people.}
These data enable users to easily formulate a complex query to a KB such as querying a list of all \textit{scientists} who are nationals of \textit{Japan}.

However, the KB also contains many entity types that are rarely useful for humans when querying a KB.
For example, \textit{Barack Obama} has four professions listed in Freebase, namely \textit{Politician}, \textit{Lawyer}, \textit{Law professor}, and \textit{Author}, but it is considered that people primarily want to retrieve Barack Obama as a \textit{Politician}.

Recently, Bast et al. \cite{DBLP:conf/sigir/BastBH15} addressed this problem by assigning a relevance score to each pair consisting of an entity and its type in KB.
These scores enable us to enhance the ranking results of entity retrieval tasks by sorting the results based on these relevance scores.

In this paper, we describe our approach for this task.
We use multiple neural network classifiers with the objective of predicting the probability of an entity type when a KB entity is given.
Notably, we introduce an attention mechanism to our neural network model in order to enable the model to prioritize a small number of relevant features.
In addition, we use another supervised machine learning model (i.e., gradient boosted regression trees (GBRT) \cite{Friedman2001}) to convert the outputs of these classifiers into the final relevance scores.

The proposed method was applied to the triple scoring task at the WSDM Cup 2017 \cite{DBLP:conf/wsdm-cup/BastBH17,DBLP:conf/wsdm/HeindorfPBBH17}
The results demonstrated that our method achieved the best results in one out of three measures (i.e., Kendall's $\tau$), and exhibited competitive performance in the other two measures (i.e., accuracy and average score difference).

\section{Our Approach}

Given a KB entity $e$ and its target type $t$, our method predicts a score that represents the relevance of $e$ belonging to $t$.
Here, we adopt a two-step approach: the first step is a classification step that aims to estimate the probability of $e$ belonging to $t$ ($P(t|e)$) using multiple neural network-based classifiers.
The second step is a scoring step that uses a supervised machine learning model to convert the outputs of these classifiers to the target relevance score.
In accordance with the task specifications for WSDM Cup 2017, our model assigns relevance scores to pairs of people and their professions, and people and their nationalities.

\subsection{Classification Step}
\label{subsec:classification-step}

We train the classifier by using all the KB entities that only have a single type, as in the previous work by Bast et al. \cite{DBLP:conf/sigir/BastBH15}.
This configuration enables us to address this problem as a multi-class classification of entities over all possible types.
It is important to note that, because our objective is assigning relevance scores to entities with multiple types, entities with only a single type can be safely used as training data.

\subsubsection{Model}

We use sets of words and entities that are relevant to $e$ as inputs to the classifier.
We adopt the neural bag-of-items model with a simple item-level attention mechanism \cite{ling-EtAl:2015:EMNLP1} to derive the representation of the set of items (i.e., words or entities).
Specifically, given a set of items, $x_1, x_2, ..., x_N$, we first compute the weighted sum of their corresponding embedding as follows:
\begin{equation}
\label{eq:weighted-sum}
\mathbf{c} = \sum_{i=1}^N a(x_i)\mathbf{v}_{x_i},
\end{equation}
Here, $\mathbf{v}_{x} \in \mathbb{R}^{d_w}$ is an embedding of $x$, and $a(x)$ is a function that computes the item-level attention weight for $x$, which is defined as the following softmax function:
\begin{equation}
\label{eq:attention}
a(x) = \frac{\exp(\mathbf{w}_a\!^\top \mathbf{u}_{x} + b_a)}{\sum_{j=1}^N\exp(\mathbf{w}_a\!^\top\mathbf{u}_{x_j} + b_a)},
\end{equation}
where $\mathbf{w}_a \in \mathbb{R}^{d_a}$ is a weight vector, $b_a \in \mathbb{R}$ is a bias, and $\mathbf{u}_x \in \mathbb{R}^{d_a}$ is an attention embedding of $x$.
The function $a(x)$ aims to capture the importance of the item $x$, thereby allowing the model to focus on a small number of relevant items.

Finally, we adopt a multi-layer perceptron (MLP) classifier with a single hidden layer with $l$ units, ReLU non-linearity, and dropout with a probability $p$.
Using Eq. \eqref{eq:weighted-sum}, we compute two feature vectors $\mathbf{c}_w$ and $\mathbf{c}_e$ using the sets of words and entities, respectively.
We then build a feature vector by concatenating $L_2$-normalized versions of vectors $\frac{\mathbf{c}_w}{||\mathbf{c}_w||}$ and $\frac{\mathbf{c}_e}{||\mathbf{c}_e||}$\footnote{We also tested the vector averaging ($\frac{\mathbf{c}}{N}$) rather than $L_2$ normalization; however, $L_2$ normalization, in general, performed marginally more accurate in terms of the classification accuracy.}, and feed the vector to MLP.

\subsubsection{Corpus}

As explained in the previous section, we train the classifier by using sets of words and entities relevant to $e$.
To extract words and entities relevant to $e$, we use the following two sources: (1) the corresponding Wikipedia articles of $e$ (denoted by \textit{article}), and (2) Wikipedia sentences that contain a link anchor that corresponds to $e$ (denoted by \textit{sentence}).
In both cases, words are extracted simply by tokenizing the text, and entities are the referent entities of link anchors in the text.
Further, in the latter case, we restrict the words to the contextual words of the link anchor in a window of length $m$.\footnote{We do not include the words within the anchor text.}

We extracted Wikipedia articles directly from the July 2016 Wikipedia dump obtained from Wikimedia Downloads\footnote{\url{https://dumps.wikimedia.org/}}.
We also used the public \textit{wiki-sentences} dataset\footnote{We downloaded the dataset from the Web site of the WSDM Cup 2017: \url{http://www.wsdm-cup-2017.org/triple-scoring.html}} to obtain the Wikipedia sentences.
In addition, we used words and entities that appear five times or more in the corpus, and simply ignored the other words and entities.

\subsubsection{Training}
\label{subsubsec:classifier-training}

All parameters used in this model were initialized randomly and updated using back-propagation.
We trained the model using stochastic gradient descent (SGD) and the learning rate was controlled by Adam \cite{kingma2014adam}.
The batch size was fixed as 100, the training consisted of one epoch, and the categorical cross-entropy was used for the loss function.
We used a NVIDIA Tesla K80 GPU to train the model.

Regarding hyper-parameters, the number of embedding dimensions $d_w$ and $d_a$ were 300 and 10, respectively; the number of units in the hidden layer $l$ was 2,000, and the dropout probability $p$ was 0.5.
We also selected the context window size of the link anchors $m$ from 5 and 10.

In addition, we optionally introduced class weights to the loss function because the distribution of the target type was highly imbalanced.
We adopted a weighted loss function based on the class weight heuristic implemented in Scikit-learn\footnote{\url{https://github.com/scikit-learn/}}.

We trained classifiers with various configurations.
Table \ref{tb:classifier-list} shows the list of configurations used to train the classifiers.
For each of the two corpora (i.e., \textit{article} and \textit{sentence}), we created eight classifiers with different training configurations, such as class weights and an attention mechanism in the enabled or disabled states\footnote{We disabled the attention mechanism by simply replacing $a(x)$ in Eq.\eqref{eq:weighted-sum} by 1.}, using either both words and entities or only entities as input, and changing the context window size.
In addition, we trained these classifiers for both the profession and nationality domains.
Therefore, the total number of classifier instances was 32.

\begin{table*}[t]
\centering
\begin{tabular}{c|c|c|c|c|c|c|c|c}
\hline
Corpus type & ID & Word & Entity & Attention & Class weight & Window & \begin{tabular}{@{}c@{}}Accuracy\\(profession)\end{tabular} & \begin{tabular}{@{}c@{}}Accuracy\\(nationality)\end{tabular}\\
\hline
\multirow{8}{*}{Article}
& 1 & \checkmark & \checkmark & \checkmark & - & - & 85.4\% & 94.7\%\\
& 2 & \checkmark & \checkmark & - & - & - & 84.5\% & 94.3\%\\
& 3 & \checkmark & \checkmark & \checkmark & \checkmark & - & 73.3\% & 91.4\%\\
& 4 & \checkmark & \checkmark & - & \checkmark & - & 70.8\% & 90.9\%\\
& 5 & - & \checkmark & \checkmark & - & - & 83.6\% & 94.3\%\\
& 6 & - & \checkmark & - & - & - & 82.5\% & 93.5\%\\
& 7 & - & \checkmark & \checkmark & \checkmark & - & 73.1\% & 90.4\%\\
& 8 & - & \checkmark & - & \checkmark & - & 70.5\% & 89.4\%\\
\hline
\multirow{8}{*}{Sentence}
& 9 & \checkmark & \checkmark & \checkmark & - & 5 & 80.6\% & 90.4\%\\
& 10 & \checkmark & \checkmark & - & - & 5 & 79.5\% & 89.2\%\\
& 11 & \checkmark & \checkmark & \checkmark & \checkmark & 5 & 56.4\% & 82.6\%\\
& 12 & \checkmark & \checkmark & - & \checkmark & 5 & 55.6\% & 80.7\%\\
& 13 & \checkmark & \checkmark & \checkmark & - & 10 & 79.0\% & 91.4\%\\
& 14 & \checkmark & \checkmark & - & - & 10 & 78.4\% & 90.3\%\\
& 15 & \checkmark & \checkmark & \checkmark & \checkmark & 10 & 55.6\% & 83.4\%\\
& 16 & \checkmark & \checkmark & - & \checkmark & 10 & 51.4\% & 81.8\%\\
\hline
\end{tabular}
\caption{Various configurations used to train the classifiers.}
\label{tb:classifier-list}
\end{table*}

\newpage
\subsection{Scoring Step}
\label{subsec:scoring-step}

We converted the outputs of the above-mentioned classifiers into relevance scores by adopting gradient boosted regression trees (GBRT) \cite{Friedman2001}.
Given an entity $e$ and a type $t$, our scoring model predicts the relevance score ranging from 0 to 7.

We experimented with two models of GBRT: the regression model and the binary classification model.
The regression model directly learns the target scores ranging from 0 to 7, whereas the binary classification model is trained using a modified dataset where the training instances with scores less than or equal to 2 are relabeled as \textit{false}, while those with scores greater than or equal to 5 are relabeled as \textit{true}, and the other instances are excluded from the training.
During the inference stage, the regression model outputs an integer value that is the closest to the estimated score. The binary classification model predicts 5 if the predicted result is true, and predicts 2 otherwise.
Moreover, we use exactly the same features for these two models.

\subsubsection{Features}

We compute the features based on two types of outputs of each classifier, the probability $P(t|e)$ and the unnormalized version of $P(t|e)$, which is the corresponding input value to the softmax layer of the MLP.
For each of the two values, we compute three features, the value itself, and the difference between the value and the minimum and the maximum value among all valid types.
It should be noted that the maximum value corresponds to the output value of the predicted type of the classifier.

Further, we observe that some pairs of types co-occur very frequently in the KB (e.g., \textit{Singer} and \textit{Singer-songwriter}).
In order to incorporate this into the model, we also use the point-wise mutual information (PMI) on the type co-occurrence data in the KB.
In particular, we add the feature representing the PMI score between the target type $t$ and the type predicted by each classifier when these two types are not equal.
Moreover, apart from the classifier outputs, we also include the number of valid types associated with $e$ in the feature set.

\subsubsection{Dataset}

We train our model by using the dataset obtained from the WSDM Cup web site.
This dataset comprises two domains, \textit{professions} and \textit{nationalities}, of person entities retrieved from Freebase.
The profession dataset and the nationality dataset contain relevance scores for 515 and 162 entity--type pairs with 134 and 77 distinct entities, respectively.
We then use this dataset for feature selection and parameter tuning as described below.

\subsubsection{Training}

We train the regression and classification models for both the profession and the nationality domains.
Feature selection is used to select a subset of the most relevant features.
We first perform a greedy forward feature selection based on the performance of 10-fold cross validation, and simply select the set of features that perform the best.
We also tune the hyper-parameters of GBRT using the selected features and the 10-fold cross validation, and use the hyper-parameters that provide the best performance.
In addition, the performance is evaluated using the mean absolute error for the regression model and the accuracy for the binary classification model.

\subsection{Implementation}

We implemented the classifier described in Section \ref{subsec:classification-step} using Python, Keras\footnote{\url{https://github.com/fchollet/keras}}, and Theano \cite{2016arXiv160502688short}.
Further, our scoring model described in Section \ref{subsec:scoring-step} was implemented using Python and Scikit-learn.
We also used Hyperopt\footnote{\url{http://hyperopt.github.io/hyperopt/}} for performing the hyper-parameter search of GBRT.

\section{Experiments}

In this section, we first describe the performance evaluation of the classifiers presented in Section \ref{subsec:classification-step}.
Then, we present the official results of the triple scoring task at the WSDM Cup 2017.

\subsection{Evaluating Classifiers}

In order to independently evaluate the performances of the proposed classifiers, we randomly selected 10\% of the KB entities with a single type, and measured the classification accuracy using these selected entities.

Table \ref{tb:classifier-list} lists the accuracies of the classifiers corresponding to various training configurations presented in Section \ref{subsubsec:classifier-training}.
As can be seen in the table, the attention mechanism effectively improved the performance, whereas the use of class weights degraded the accuracy in general.
Further, the classifiers trained with the article corpus generally performed more accurately than those trained with the sentence corpus.

We also found in our experiments that incorporating the outputs of classifiers that achieve lower accuracies often improved the performance of the scorer.
Therefore, the strategy we adopted used the outputs of various classifiers rather than focusing on the outputs of a single accurate classifier.

Further, in order to investigate how the attention model works in practice, we inspected the words and entities having large attention weights $\mathbf{w}_a\!^\top \mathbf{u}_{x}$ in Eq.\eqref{eq:attention}.
Table \ref{tb:top-attention-words} and Table \ref{tb:top-attention-entities} presents the top 10 words and entities with large weights, respectively.
These weights were extracted from classifier 1, which was trained for the profession domain.
It appeared that our classifier effectively focused on words and entities that strongly indicate a profession.
For example, the top words included various professions, such as \textit{physicists} and \textit{economists}, and all the top entities were lists or categories that were strongly associated with a profession.

\begin{table}[t]
\centering
\begin{tabular}{c|p{3cm}}
\hline
Rank & Top words\\
\hline
1 &
physicists \\
2 &
economists \\
3 &
mathematicians \\
4 &
psychologists \\
5 &
draftexpress \\
6 &
novelists \\
7 &
bàsquet \\
8 &
botanists \\
9 &
aoni \\
10 &
barristers \\
\hline
\end{tabular}
\caption{Top 10 words with large attention weights.}
\label{tb:top-attention-words}
\end{table}

\begin{table*}[t]
\centering
\begin{tabular}{c|p{10.5cm}}
\hline
Rank & Top entities\\
\hline
1 &
Category:Members of the United States House of Representatives from New York\\
2 &
List of Major League Baseball career stolen bases leaders\\
3 &
Category:Liberal Party of Australia members of the Parliament of Australia\\
4 &
Category:Shooters at the 2012 Summer Olympics\\
5 &
Category:American science writers\\
6 &
Category:National Hockey League first round draft picks\\
7 &
Category:Cleveland Browns players\\
8 &
Category:American anthropologists\\
9 &
List of drummers\\
10 &
Category:Tennessee Titans players\\
\hline
\end{tabular}
\caption{Top 10 entities with large attention weights.}
\label{tb:top-attention-entities}
\end{table*}

\newpage
\subsection{Competition Results}

We submitted our proposed method to the triple scoring task at the WSDM Cup 2017.
In this competition, the submitted methods were evaluated based on the following three measures:

\begin{itemize}
\item \textbf{Accuracy}, which is the percentage for which the estimated score differs from the score from the ground truth by at most 2.
\item \textbf{Average score difference}, which is the average score difference between the estimated scores and the ground truth scores.
\item \textbf{Kendall's $\tau$}, which is the average Kendall's $\tau$ score\footnote{Following Bast et al. \cite{DBLP:conf/sigir/BastBH15}, we used the modified version of Kendall's $\tau$ score proposed in Fagin et al. \cite{Fagin2004}} between the estimated scores and the ground truth scores.
The $\tau$ score is computed for each entity, and the final score is averaged over all entities.
\end{itemize}

Experiments were conducted using the 710 entity--type pairs containing the instances of 513 profession pairs and 197 nationality pairs.
We used different scoring models trained with the corresponding dataset for each domain.
Note that the accuracy described here is different from the accuracy used to evaluate the classifiers in the previous section.

Table \ref{tb:experimental-results} contains the official results of our methods based on the regression model (reg) and the binary classification model (clf) compared with the other top five methods proposed by competitors in terms of accuracy.
The table lists the accuracies (acc), the average score differences (asd), and the Kendall's $\tau$ scores (tau).

Our regression model achieved the best performance in terms of Kendall's $\tau$ scores among all the methods, and performed competitively in the accuracy and the average score difference.
Further, the performance of our binary classification model was superior, particularly in terms of accuracy.

\begin{table}[t]
\centering
\begin{tabular}{p{2.3cm}|c|c|c}
\hline
Name & Acc & Asd & Tau\\
\hline
Our method (reg) & 0.77 & 1.59 & \textbf{0.29}\\
Our method (clf) & 0.82 & 1.76 & 0.36\\
\hline
bokchoy1  & \textbf{0.87} & 1.63 & 0.33\\
bokchoy2  & 0.82 & \textbf{1.50} & 0.32\\
radicchio & 0.80 & 1.69 & 0.40\\
catsear   & 0.80 & 1.86 & 0.41\\
cress     & 0.78 & 1.61 & 0.32\\
\hline
\end{tabular}
\caption{Experimental results of our methods compared with the other top five methods submitted to WSDM Cup 2017.}
\label{tb:experimental-results}
\end{table}

\section{Conclusions}

In this study, we proposed an approach for assigning a relevance score to each entity--type pair in a given KB.
We trained neural network-based multiple classifiers by directly using the KB data, and converted the results of these classifiers into target relevance scores using a supervised machine learning model (i.e., GBRT).
It is worth noting that the item-based attention model we introduced to the neural network model had not been applied to this kind of task previously.
The experimental results confirmed the superiority of our approach;
we achieved the best performances in terms of Kendall's $\tau$ scores, and performed competitively in terms of the accuracy and average score difference.
We publicized the source code of our proposed method at \textsf{\url{https://github.com/wsdm-cup-2017/lettuce}} to enable it to be used for further academic research.

{\raggedright
\bibliography{library}}

\begin{thebibliography}{11}
\providecommand{\natexlab}[1]{#1}
\providecommand{\url}[1]{\texttt{#1}}
\expandafter\ifx\csname urlstyle\endcsname\relax
  \providecommand{\doi}[1]{doi: #1}\else
  \providecommand{\doi}{doi: \begingroup \urlstyle{rm}\Url}\fi

\bibitem[Auer et~al.(2007)Auer, Bizer, Kobilarov, Lehmann, Cyganiak, and
  Ives]{Auer2007}
S.~Auer, C.~Bizer, G.~Kobilarov, J.~Lehmann, R.~Cyganiak, and Z.~Ives.
\newblock {DBpedia: A Nucleus for a Web of Open Data}.
\newblock \emph{The Semantic Web}, pages 722--735, 2007.

\bibitem[Bast et~al.(2015)Bast, Buchhold, and
  Haussmann]{DBLP:conf/sigir/BastBH15}
H.~Bast, B.~Buchhold, and E.~Haussmann.
\newblock Relevance scores for triples from type-like relations.
\newblock In \emph{{SIGIR}}, pages 243--252. {ACM}, 2015.

\bibitem[Bast et~al.(2017)Bast, Buchhold, and
  Haussmann]{DBLP:conf/wsdm-cup/BastBH17}
H.~Bast, B.~Buchhold, and E.~Haussmann.
\newblock {Overview of the Triple Scoring Task at the WSDM Cup 2017}.
\newblock In \emph{WSDM Cup}, 2017.

\bibitem[Bollacker et~al.(2008)Bollacker, Evans, Paritosh, Sturge, and
  Taylor]{Bollacker2008}
K.~Bollacker, C.~Evans, P.~Paritosh, T.~Sturge, and J.~Taylor.
\newblock {Freebase: A Collaboratively Created Graph Database for Structuring
  Human Knowledge}.
\newblock In \emph{SIGMOD}, pages 1247--1250, 2008.

\bibitem[Fagin et~al.(2004)Fagin, Kumar, Mahdian, Sivakumar, and
  Vee]{Fagin2004}
R.~Fagin, R.~Kumar, M.~Mahdian, D.~Sivakumar, and E.~Vee.
\newblock {Comparing and Aggregating Rankings with Ties}.
\newblock In \emph{PODS}, page~47, 2004.

\bibitem[Friedman(2001)]{Friedman2001}
J.~H. Friedman.
\newblock {Greedy Function Approximation: A Gradient Boosting Machine}.
\newblock \emph{The Annals of Statistics}, 29\penalty0 (5):\penalty0
  1189--1232, 2001.

\bibitem[Heindorf et~al.(2017)Heindorf, Potthast, Bast, Buchhold, and
  Haussmann]{DBLP:conf/wsdm/HeindorfPBBH17}
S.~Heindorf, M.~Potthast, H.~Bast, B.~Buchhold, and E.~Haussmann.
\newblock {WSDM Cup 2017: Vandalism Detection and Triple Scoring}.
\newblock In \emph{{WSDM}}. ACM, 2017.

\bibitem[Kingma and Ba(2014)]{kingma2014adam}
D.~Kingma and J.~Ba.
\newblock {Adam: A Method for Stochastic Optimization}.
\newblock \emph{arXiv preprint arXiv:1412.6980}, 2014.

\bibitem[Ling et~al.(2015)Ling, Tsvetkov, Amir, Fermandez, Dyer, Black,
  Trancoso, and Lin]{ling-EtAl:2015:EMNLP1}
W.~Ling, Y.~Tsvetkov, S.~Amir, R.~Fermandez, C.~Dyer, A.~W. Black, I.~Trancoso,
  and C.-C. Lin.
\newblock {Not All Contexts Are Created Equal: Better Word Representations with
  Variable Attention}.
\newblock In \emph{EMNLP}, pages 1367--1372, 2015.

\bibitem[{Theano Development Team}(2016)]{2016arXiv160502688short}
{Theano Development Team}.
\newblock {Theano: A Python Framework for Fast Computation of Mathematical
  Expressions}.
\newblock \emph{arXiv preprint arXiv:1605.02688}, 2016.

\bibitem[Vrande{\v{c}}i{\'{c}} and Kr{\"{o}}tzsch(2014)]{Vrandecic2014a}
D.~Vrande{\v{c}}i{\'{c}} and M.~Kr{\"{o}}tzsch.
\newblock {Wikidata: A Free Collaborative Knowledgebase}.
\newblock \emph{Communications of the ACM}, 57\penalty0 (10):\penalty0 78--85,
  2014.

\end{thebibliography}
\end{document}